\documentclass{article}
\pdfoutput=1
\usepackage[final]{nips_2018}

\usepackage[utf8]{inputenc} 
\usepackage[T1]{fontenc}    
\usepackage{hyperref}       
\usepackage{url}            
\usepackage{booktabs}       
\usepackage{amsfonts}       
\usepackage{nicefrac}       
\usepackage{microtype}      
\usepackage{courier}
\usepackage{graphicx}
\graphicspath{ {/} }
\usepackage{amsmath}
\usepackage{amssymb}
\usepackage{wrapfig}
\usepackage{multirow}
\usepackage{bbm}
\usepackage{bm}

\usepackage{subcaption}

\newcommand{\norm}[1]{\|#1\|}
\newcommand{\given}{\: | \:}
\newcommand{\defined}{\triangleq}

\title{ \Large Joint Visual-Textual Embedding for Multimodal Style Search}
\author{
	Gil~Sadeh\\
	Amazon Lab126 \\
	{\scriptsize \texttt{gilsadeh@amazon.com} }\\
	\And
	Lior~Fritz\\
	Amazon Lab126 \\
	{\scriptsize \texttt{liorf@amazon.com}} \\
	\And
	Gabi~Shalev\\
	Amazon Lab126 \\
	{\scriptsize \texttt{shalevg@amazon.com}} \\
	\And
	Eduard~Oks\\
	Amazon Lab126 \\
	{\scriptsize \texttt{oksed@amazon.com}} \\
}

\begin{document}

\maketitle

\begin{abstract}

We introduce a multimodal visual-textual search refinement method for fashion garments. 
Existing search engines do not enable intuitive, interactive, refinement of retrieved results based on the properties of a particular product. 
We propose a method to retrieve similar items, based on a query item image and textual refinement properties. We believe this method can be leveraged to solve many real-life customer scenarios, in which a similar item in a different color, pattern, length or style is desired. We employ a joint embedding training scheme in which product images and their catalog textual metadata are mapped closely in a shared space. This joint visual-textual embedding space enables manipulating catalog images semantically, based on textual refinement requirements. We propose a new training objective function, Mini-Batch Match Retrieval, and demonstrate its superiority over the commonly used triplet loss. Additionally, we demonstrate the feasibility of adding an attribute extraction module, trained on the same catalog data, and demonstrate how to integrate it within the multimodal search to boost its performance. We introduce an evaluation protocol with an associated benchmark, and compare several approaches.

\end{abstract}

\section{Introduction}

Recently, the ability to embed representations of images and text in a joint space was studied thoroughly for many tasks. Among which are image annotation and search~\cite{klein2015associating, chen2017amc}, zero-shot recognition~\cite{socher2013zero, mukherjee2016gaussian, frome2013devise, reed2016learning}, robust image classification~\cite{frome2013devise}, image description generation~\cite{karpathy2015deep}, visual question-answering~\cite{nam2017dual} and more.

Vector arithmetic properties have been demonstrated lately as a surprising artifact of learning semantic embedding spaces. Mikolov et al.~\cite{mikolov2013linguistic} showed that a learned \emph{word2vec} embedding space can capture semantic vector arithmetics, such as: ``Paris'' - ``France'' +``Italy'' = ``Rome''. Kiros et al.~\cite{uniVSE} demonstrated a similar phenomenon in multimodal visual-semantic embedding spaces, in which, with linear encoders, the learned embedding space captures multimodal regularities. For instance, given $ f_I $, a representing vector of an image of a blue car, $ f_I $ - ``blue'' + ``red'' yields a representing vector of a red car image.

\begin{figure}[t]
	\centering
	\includegraphics[width=0.75\textwidth]{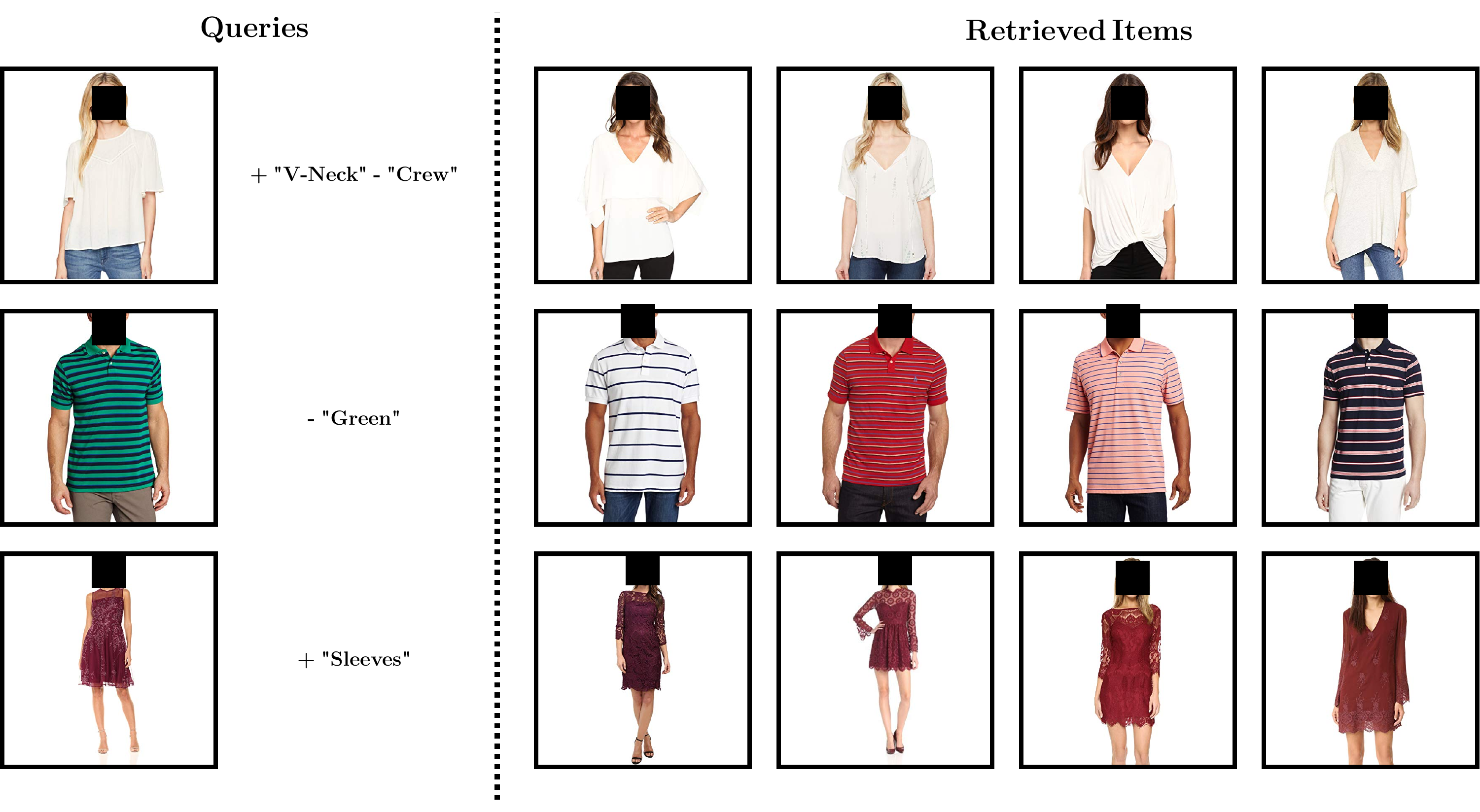}
	\caption{Examples of typical multimodal search queries and their top retrieved results.}
	\label{fig:MMS-examples}
\vspace*{-10pt}
\end{figure}

This paper refers to the specific, fine-grained, task of visual-textual multimodal search in the fashion domain. Example queries and their retrieved results can be seen in Figure~\ref{fig:MMS-examples}.
We believe this type of application can greatly impact the customer shopping experience, by enabling intuitive and interactive search refinements. 
With this technology, browsing large fashion catalogs and finding specific products can become easier and less frustrating. 

We consider training a visual-textual joint embedding model in an end-to-end manner, based on images and textual metadata of catalog products. We propose a training objective function which we refer to as Mini-Batch Match Retrieval (MBMR). Each mini-batch consists of matching and non-matching image-text pairs. We compute the cosine similarity of each pair, and maximize matching samples similarities with cross-entropy loss, as done in~\cite{wojke2018deep}. However, unlike ~\cite{wojke2018deep}, which assigns an embedding vector per each category, in our retrieval task the notion of category does not exist. Instead, we learn an embedding for each item (image and text) and try to classify the correct pair from the mini-batch reference set.
We demonstrate the superiority of this approach over the commonly used triplet loss.

In addition, we explore the task of visual fashion attribute extraction, utilizing the noisy catalog data alone, without additional annotation effort. A pool of possible fashion attributes is extracted from frequent words in the catalog metadata, and a multi-label classifier is trained to extract the correct ones given a product image. We demonstrate that, although the catalog-based labels are noisy, attribute extraction produces satisfying results. 

We propose and evaluate several approaches for multimodal search. The first approach leverages the query-arithmetic phenomenon of visual-textual joint embeddings. A second approach utilizes our learned attribute extraction module, for soft textual filtering, alongside visual search, based on our visual-semantic embedding space. Finally, we propose a combined approach, which leverages both the joint embedding based query-arithmetic property and soft attribute filtering. This approach yields a considerable performance improvement over the other methods.

\section{Related Work}

Image recognition classifiers treat labels as disconnected and unrelated, resulting in visual recognition systems that cannot transfer semantic information about learned labels to unseen words or phrases. Early visual-textual joint embedding works addressed this problem by mapping image-word pairs, where the words corresponded to image labels or attributes. Weston et al.~\cite{weston2010large} trained a joint embedding model of both images and their labels, by employing an online learning-to-rank algorithm. Frome et al.~\cite{frome2013devise} leveraged textual data to learn semantic relationships between labels by explicitly mapping images into a common visual-semantic embedding space. They showed this approach leads to more semantically reasonable errors and significantly improved zero-shot predictions.

More recent works attempted to map images and their textual descriptions into a common embedding space.
Klein et al.~\cite{klein2015associating} employed Canonical Correlation Analysis (CCA)~\cite{cca} to learn projections of precomputed image and caption features, onto a joint embedding space, for cross-modal retrieval tasks. Kiros et al.~\cite{uniVSE} employed a triplet-based ranking loss in order to learn a similar embedding space for images and text, for caption generation and ranking tasks. 
Karpathy et al.\cite{karpathy2014deep} worked on a finer level, embedding fragments of images and sentences jointly with a max-margin based objective. In the fashion domain, Han et al. ~\cite{han2017learning} learned a similar visual-semantic embedding for product images and their corresponding textual descriptions. 
They combined this joint embedding in their outfit recommendation engine, so that it is agnostic to the input type (image, text or a combination of both).

Several works considered the task of manipulating attributes for fashion search. 
Zhao et al.~\cite{zhao2017memory} trained a network to jointly optimize attribute classification loss and triplet ranking loss, over image triplets, for facilitating precise attribute manipulation and image retrieving. The network learned, in a supervised manner, to modify the intermediate image representation based on the desired manipulation.
Kenan et al.~\cite{ak2018fashionsearchnet} proposed learning attribute specific representations by leveraging weakly-supervised localization, in order to manipulate attributes during fashion search.
M. G{\"u}nel et al.~\cite{gunel2018language} proposed a GAN-based solution for language guided image manipulation, where the generator performs feature-wise linear modulation between visual features and desired natural language descriptions.

Zhao et al. ~\cite{zhao2018multi} proposed a Multi-Task Learning (MTL) system to jointly train an image captioning and attribute extraction model. They demonstrated how the auxiliary attribute extraction task resulted in better image representation and improved performance in the original captioning task.

\section{Data}
The data used for training the joint embedding model consists of 0.5M fashion products from a retail website. Each product item has associated image and textual metadata. The catalog has a very diverse range of products, and includes rich and relatively accurate metadata. The actual search, and its evaluation, are performed on an larger set of 1.5M catalog items (only tops, bottoms and dresses) from a different retail website.

Although the textual metadata of our training catalog is relatively clean and accurate compared to other catalogs, there still exists noise and variability in the textual metadata. Similar items can have very different textual descriptions, while non-similar items may have relatively similar descriptions. 
Moreover, textual metadata may be lacking in details frequently.

\begin{figure}[b]

	\centering
	\includegraphics[width=1.0\textwidth]{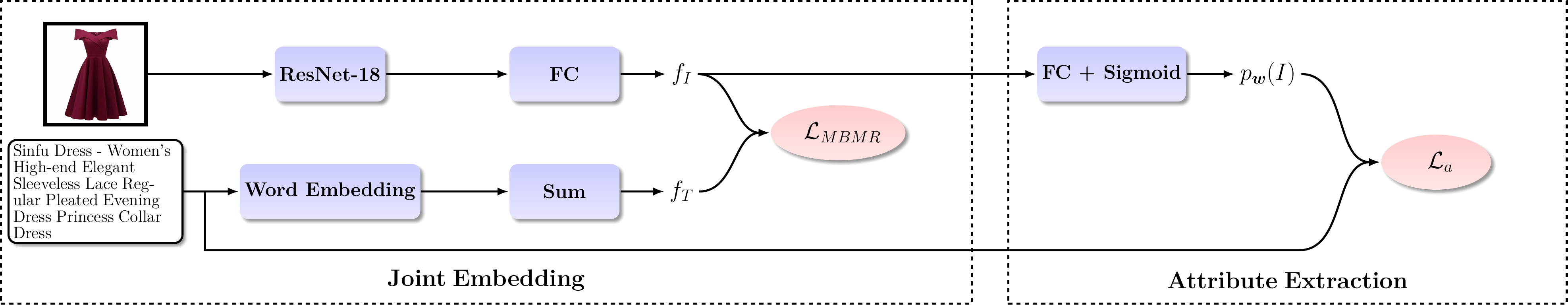}
	\caption{ \textbf{Joint Embedding:} A ResNet-18 CNN extracts visual features from the image with an additional fully connected (FC) layer which projects these features to the joint space. The textual encoder sums the word embeddings of all relevant words in the textual metadata. \textbf{Attribute Extraction:} An additional network branch extracts attribute probabilities from the image representation. It utilizes the catalog textual metadata as ground-truth attribute labels. }
	\label{fig:model}

	\vspace*{-10pt}
\end{figure}

\section{Training}
\label{sec:training}

Figure~\ref{fig:model} illustrates the architecture of our model. 
The basic joint embedding model consists of two main branches, an image encoder and a text encoder. 
Image encoding is based on a ResNet-18~\cite{he2016deep} deep convolutional neural network (CNN), followed by an additional fully connected layer which projects the visual feature vector to the same space as the textual encoding. 
Text encoding is done by summing the word embeddings of all input words. 
The text is treated as a bag-of-words, rather than an ordered sequence of words, since it is accumulated from several metadata fields, and may contain a mixture of sentences and individual keywords.

For attribute extraction, we add a third branch to this joint embedding architecture. The branch consists of a fully connected layer, followed by a sigmoid activation function for multi-label classification. The input to this branch is the image feature vector, $f_I$, and the output is a vector of attribute probabilities, $p_{\bm w}(I)$. 
The size of the attribute probability vector is determined by the vocabulary size, $|V|$.

The model is trained end-to-end. That is, both encoder branches are trained jointly. 
The ResNet weights are initialized by a pre-trained ImageNet~\cite{deng2009imagenet} model. The word embeddings are based on \emph{word2vec}, and are trained on product titles. Word embeddings that do not appear in this set are initialized randomly. The fully connected layer parameters are initialized with PCA over the extracted ImageNet features. We also fix the ResNet weights at the begining of training, and unfreeze only the two top Residual blocks after two epochs. We use the Adam~\cite{kingma2014adam} optimizer, with an exponentially decaying learning rate schedule. We have found that all of these settings are helpful in order to improve convergence and reduce overfitting.

Our training objective is composed of two loss terms. 
A Mini-Batch Match Retrieval (MBMR) loss, $\mathcal{L}_{MBMR}$, for the task of learning a joint embedding space, and a multi-label cross-entropy loss, $\mathcal{L}_a$, for attribute extraction. The final objective is a weighted sum of both loss terms.

\subsection{Textual Metadata Preprocessing}
In order to clean and normalize the textual metadata we use several preprocessing steps when building our vocabulary. (1) Tokenization – divide the raw description text into a set of tokens.
(2)	Stemming – normalize words to their base form, in order to avoid multiple word variations with the same visual meaning.
(3) Part-Of-Speech (POS) based filtering – identify noun and adjective tokens, which are more likely to have visual significance, and ignore the rest.
(4) Word frequency thresholding – words that appear less times in the dataset than some hard cut-off threshold are removed, thus reducing noise and avoiding an unnecessarily large vocabulary. We set our threshold to $500$.

These preprocessing steps determine the vocabulary, $V$, of our model. Its size, $|V|$, also affects the number of parameters in the word embeddings and attribute extraction fully connected layer.

\subsection{Mini-Batch Match Retrieval Objective}
\label{mbmr}

The objective of the joint-embedding training procedure should encourage matching (non-matching) image-text pairs to be as close (distant) as possible to (from) each other, in the common embedding space. To achieve this, we propose the following Mini-Batch Match Retrieval (MBMR) objective.  

In our training setting, each mini-batch consists of $N$ product items, $ \left\{I_i, T_i \right\}_{i=1}^{N} $, where $ I_i $ is an image, and $ T_i $ is its corresponding textual metadata.
For each image embedding in the batch, $ f_I $, and text embedding in the batch, $ f_T $, we compute their cosine similarity,
\begin{equation}
S_{I,T} = \dfrac {f_I \cdot f_T} {\norm{f_I} \norm{f_T}}.
\end{equation}

We then define the probability of image $ I_i $ to match description $ T_j $ as,
\begin{equation}
P(T_j \given I_i) = \dfrac{\exp\left\{S_{I_i,T_j} / \tau\right\}}{\sum_k{\exp\left\{S_{I_i,T_k} / \tau \right\}}},
\end{equation}

where $ \tau $ is a temperature parameter. The probability of $ T_i $ to match image $ I_j $, is calculated similarly,

\begin{equation}
P(I_j \given T_i) = \dfrac{\exp\left\{S_{I_j, T_i} / \tau\right\}}{\sum_k{\exp\left\{S_{I_k, T_i } / \tau \right\}}}.
\end{equation}

The final objective is obtained by applying cross-entropy for every query image and text in the batch,
\begin{equation}
\mathcal{L}_{MBMR} = - \sum_i{\log P(T_i \given I_i)}  - \sum_i{\log P(I_i \given T_i)}.
\end{equation}
\vspace{-20pt}

\subsection{Attributes Extraction}

Since our model learns to bridge the gap between images and text, it is natural to expect it to be able to provide out-of-the-box attribute extraction just by computing cosine-similarities between images and words. In practice, however, this leads to noisy results, due to the following reasons. 
First, not all words are equally visually grounded. Some words are very visually dominant, while others may have very little (if any) visual significance, and may exist only due to imperfect textual preprocessing.
Second, word frequencies vary significantly. Some attributes appear in almost every item description in the catalog, like garment types, while others appear very rarely. This data behavior can be considered as noisy labels for our attribute extractor.

In order to create a more robust attribute extraction model, we add another branch to the model which consists of a fully connected layer that projects image embeddings to the vocabulary size, $|V|$, followed by a sigmoid function. The outputs of this branch, $ \left\{\hat p_w(I)\right\} $, are approximations of the probabilities for each word $ w $ in the vocabulary to belong to image $ I $. The ground-truth labels are determined by the existence of words in the product textual metadata. An additional loss term is added for this multi-label classification task.

During inference, we take the following additional steps in order to obtain reliable attribute extraction. We compute a per-word threshold, by optimizing the F-score on the validation set. This threshold, $thr_w$, is used to define a classification score,

\begin{equation}
\tilde{p}_w(I) = {\rm sigmoid}\left( \dfrac{\hat p_w (I)-{\rm thr}_w}{{\rm thr}_w }\right).
\end{equation}

Additionally, we compute a cosine-similarity score between word and image features,
\begin{equation}
 S_{w, I} = \dfrac {f_w \cdot f_I} {\norm{f_w} \norm{f_I}},
\end{equation}

where $f_w$ and $f_I$ are the word and image embeddings, in the joint space, respectively.

Finally, we average the classification score, $\tilde{p}_w(I)$, and the clipped cosine-similarity score, $S_{w, I}$, in order to obtain the final probability that word $ w $ is a characteristic of image $ I $,

\begin{equation}
p_w(I) =  \dfrac{\tilde{p}_w(I)  + \max{(S_{w, I},0)}}{2}.
\end{equation}

In order to approximate the probability of a desired and undesired attribute set  $\bm w= \left\{\bm w^+, \bm w^-\right\}$, in the multimodal search scenario,  we follow Bayes rule, under the independence assumption,

\begin{equation}
p_{\bm w}(I) = \prod_{w^+ \in \bm w^+}{p_{w^+}(I)} \prod_{w^- \in \bm w^-}{(1-p_{w^-}(I))}.
\end{equation}

\begin{figure}[t]
	\centering
	\label{fig:attribute-examples}
	\includegraphics[width=0.65\textwidth]{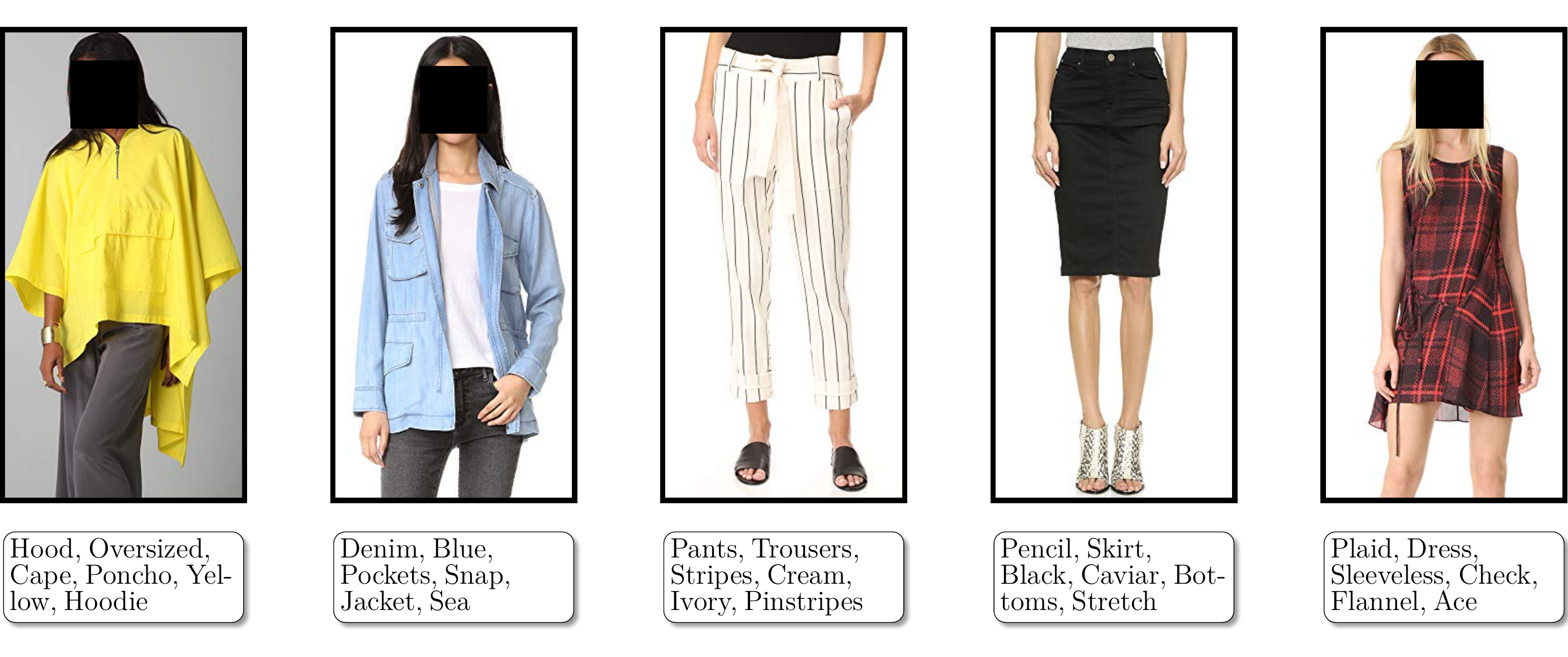}
	\caption{Attributes extraction examples. We list the top 6 extracted attributes, for each catalog image, according to their probabilities $p_w(I)$. }

\end{figure}

\section{Multimodal Refinement Search}

\subsection{Query Arithmetic Approach}
\label{sec:query_arithemtic}
During inference, the text and image encoders can yield image and textual query feature vectors which lay in a common embedding space. These feature vectors can be used to search for products, with similar visual or textual properties, in a dedicated catalog. The similarity metric used for matching the query and catalog items is, as in the training phase, cosine similarity. The catalog image and textual features can be precomputed offline once.

Ideally speaking, the fact that visual and textual modalities share the same embedding space, combined with the linear nature of the text encoder,  enables performing arithmetic operations (as in \emph{word2vec}) in order to manipulate the desired search query. This enables searching for visually similar products with some different properties, defined textually, by simply adding (subtracting) desired (undesired) textual features to (from) the product visual feature vector. That is, for a given query image, $I$, and a desired and undesired attribute set, $\bm w= \left\{\bm w^+, \bm w^-\right\}$, the new mutlimodal query $ q $ can be defined by,

\begin{equation}
q = f_I + f_T,
\end{equation}
\begin{equation}
f_T = \sum_{w^+ \in \bm w^+}{f_{w^+}} -  \sum_{w^- \in \bm w^-}{f_{w^-}},
\end{equation}

where $ f_I $ is the image embedding, and $ f_T $ is the linear combination of desired and undesired word embeddings.

The similarity score, $ S $, between the query and reference catalog items, is defined as the cosine similarity between  $q$ and the reference visual features $f_{I_r}$. 

\subsection{Attribute Filtering Approach}
\label{sec:attribute_filtering}
An alternative approach for multimodal search is filtering out all catalog items which are not consistent with the textual query. Then, the search score can be calculated based on visual similarity alone. This approach can be formulated as follows.
\begin{equation}
S =  \dfrac {q \cdot f_{I_r}} {\norm{q} \norm{f_{I_r}}} \cdot \mathbbm{1}{( w \in T_r \;\;\; \forall w \in \bm w^+)} \cdot \mathbbm{1}{( w \notin T_r \;\;\; \forall w \in \bm w^- )},
\end{equation}
where $q = f_I $, $ T_r $ is the set of words in the reference textual metadata, and $ \bm w^+ $ ($ \bm w^-$ ) is the set of desired (undesired) properties.

This approach should work well given an ideal catalog, with complete and error-free textual metadata. 
However, this is not the case in most catalogs. Hence, we derive a soft filtering method based on attribute extraction probabilities,

\begin{equation}
S =  \dfrac {q \cdot f_{I_r}} {\norm{q} \norm{f_{I_r}}} \cdot p_{\bm w} (I_r),
\end{equation}
where $q = f_I$ and $p_{\bm w} (I_r)$ is the probability of the textual desired and undesired properties in the reference image $I_r$.

\subsection{Combined Approach}
We attempt to combine both previously described methods into a single robust one. We do so by using the soft attribute filtering along with the query arithmetic based search.
The motivation of incorporating attribute filtering is to better meet the textual manipulation criteria. Since attribute filtering is soft and noisy, it is not enough to use it with visual search alone (as in \ref{sec:attribute_filtering}), as it will encourage retrieval of visually similar items without considering the textual manipulation.

The exact formulation is as follows.

\begin{equation}
S =\dfrac {q \cdot f_{I_r}} {\norm{q} \norm{f_{I_r}}}  \cdot p_{\bm w}(I_r),
\end{equation}
where $q = f_I + f_T$, as in \ref{sec:query_arithemtic}.

\section{Evaluation}
\label{eval}

\begin{figure}[b]
	
	\includegraphics[width=0.7\textwidth]{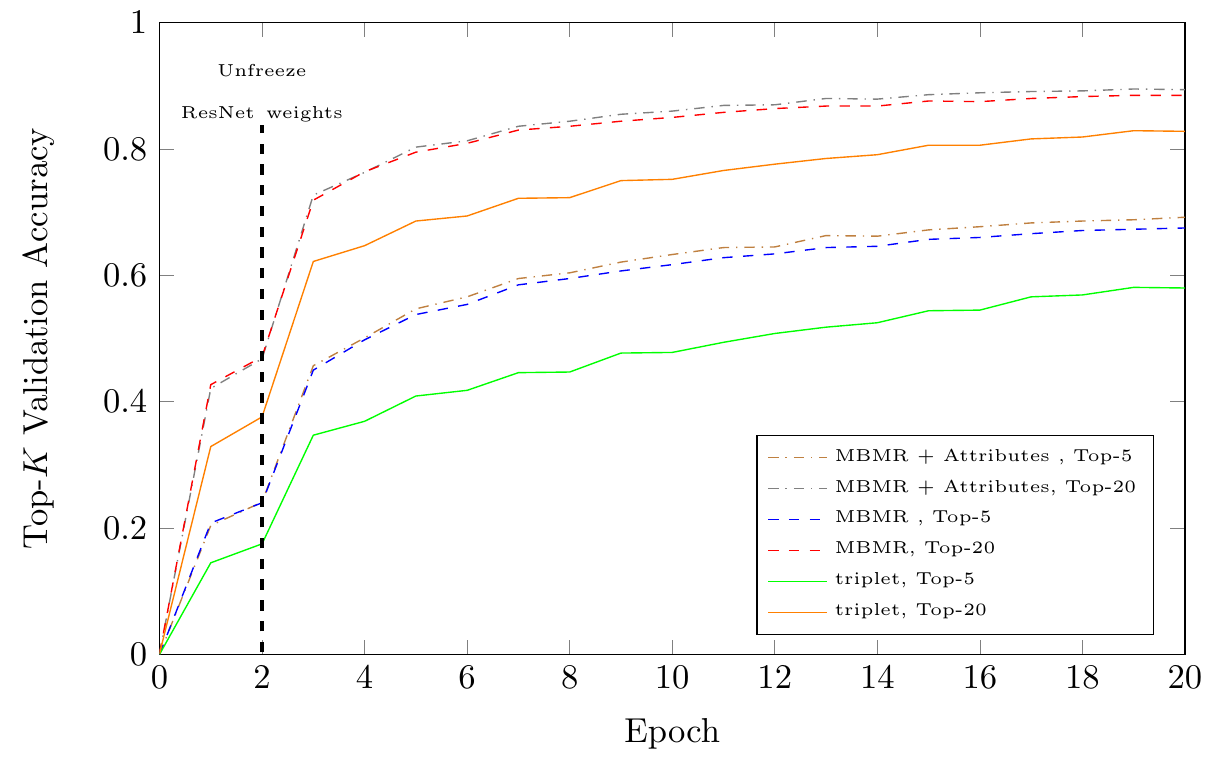}
	\centering

	\caption{ Comparison of top-$K$ validation accuracy convergence, during training, between triplet loss, MBMR loss and a multi-task objective composed of MBMR loss and multi-label cross-entropy loss for attribute extraction. 
	}
	\label{fig:mbmr_exp}

\end{figure}

For evaluation purposes we automatically constructed a benchmark of multimodal queries. Query product images (of tops, bottoms and dresses) were randomly sampled, and assigned with desired and undesired textual requirements out of a pool of common fashion attributes. The pool consisted of $110$ fashion attributes from $5$ major categories: color, pattern, neckline, style and garment type. Textual requirements can specify either adding, removing or replacing specific properties to or from the query image. The final benchmark consists of 1500 queries (300 for each attribute category), after manual verification and query filtering.

A commonly used metric in information retrieval tasks is the normalized Discounted Cumulative Gain (nDCG) \cite{jarvelin2002cumulated}. The DCG metric measures ranking quality, which cumulates the relevance of the top-$K$ retrieved items per query, while penalizing them differently based on their rank.
\begin{equation}
{\rm DCG_K} = \sum_{i=1}^{K} \dfrac{rel_i}{log_2{(i+1)}},
\end{equation}

where $ rel_i $ is the relevance of the reference item ranked in place $ i $ by the model. The relevance is given by some oracle.
The nDCG normalizes the DCG metric by the Ideal-DCG value (IDCG), which is calculated similarly to the DCG, over an ideally sorted reference list. For IDCG, we used an upper-bound approximation which assumes our reference corpus contains $K$ items with $rel = 1$.

In order to evaluate multimodel search performance, two aspects need to be accounted for, visual and textual. A perfect result would meet the textual criteria while still being as visually similar as possible to the query image. We develop two nDCG metrics, with relevance scores based on a visual oracle and a textual oracle. Our final, multimodal, metric is a simple geometric mean of both nDCG scores.

\begin{itemize}
	\itemsep-0.16em
	\item \textbf{Visual nDCG (V-nDCG)}: Based on visual relevance, which is extracted from a baseline visual search model. This purely visual model was trained with triplet loss on catalog images, where for each query image a different image of the same item was considered as a positive sample and images of different items were considered as negative samples. The relevance is the cosine similarity between reference and query visual features, extracted from this baseline model.
	\item \textbf{Textual nDCG (T-nDCG)}: Based on presence (absence) of desired (undesired) query words in the reference textual metadata. The relevance is defined by the rate of criteria that are met. A desired word criterion is considered as met if the reference metadata includes the word. An undesired word criterion is met if the reference metadata does not include the word.
	\item \textbf{Multimodal (MM)}: 
	\mbox{\rm{MM}$ \defined  \sqrt{\text{V-nDCG} \cdot  \text{T-nDCG}}$}.
\end{itemize}

These metrics are somewhat noisy, and may be inaccurate in specific cases, such as incomplete and inaccurate metadata or inaccuracies caused by the baseline visual search model. However, on the corpus level we observe that they are stable and reliable enough to serve as evaluation metrics, and help us compare different methods.

\section{Experimental Results}
\label{sec:exp}

We compare our described Mini-Batch Match Retrieval (MBMR) objective with a triplet loss, as utilized in~\cite{uniVSE}. Figure \ref{fig:mbmr_exp} shows the convergence of top-5 and top-20 validation accuracy during the joint embedding training procedure. The top-$ K $ accuracy metric measures the rate of images and text descriptions for which the actual matching pair was ranked, based on cosine similarity, within the top $K$ references out of the entire validation set, which consists of 23.5K items. In our experiments, the mini-batch size was set to $160$, the MBMR temperature $\tau$ to 0.025 and the triplet loss margin to 0.2. We believe that top-$ K $ accuracy is a good metric for this task, as in retrieval tasks we usually mostly care about the top retrieved results. It can be seen that the MBMR objective leads to faster and superior convergence over triplet loss. Additionally, it can be seen that multi-task training, with the additional attribute extraction branch and corresponding loss, slightly increases performance.

\begin{figure}[h]
	\centering
	\includegraphics[width=1\textwidth]{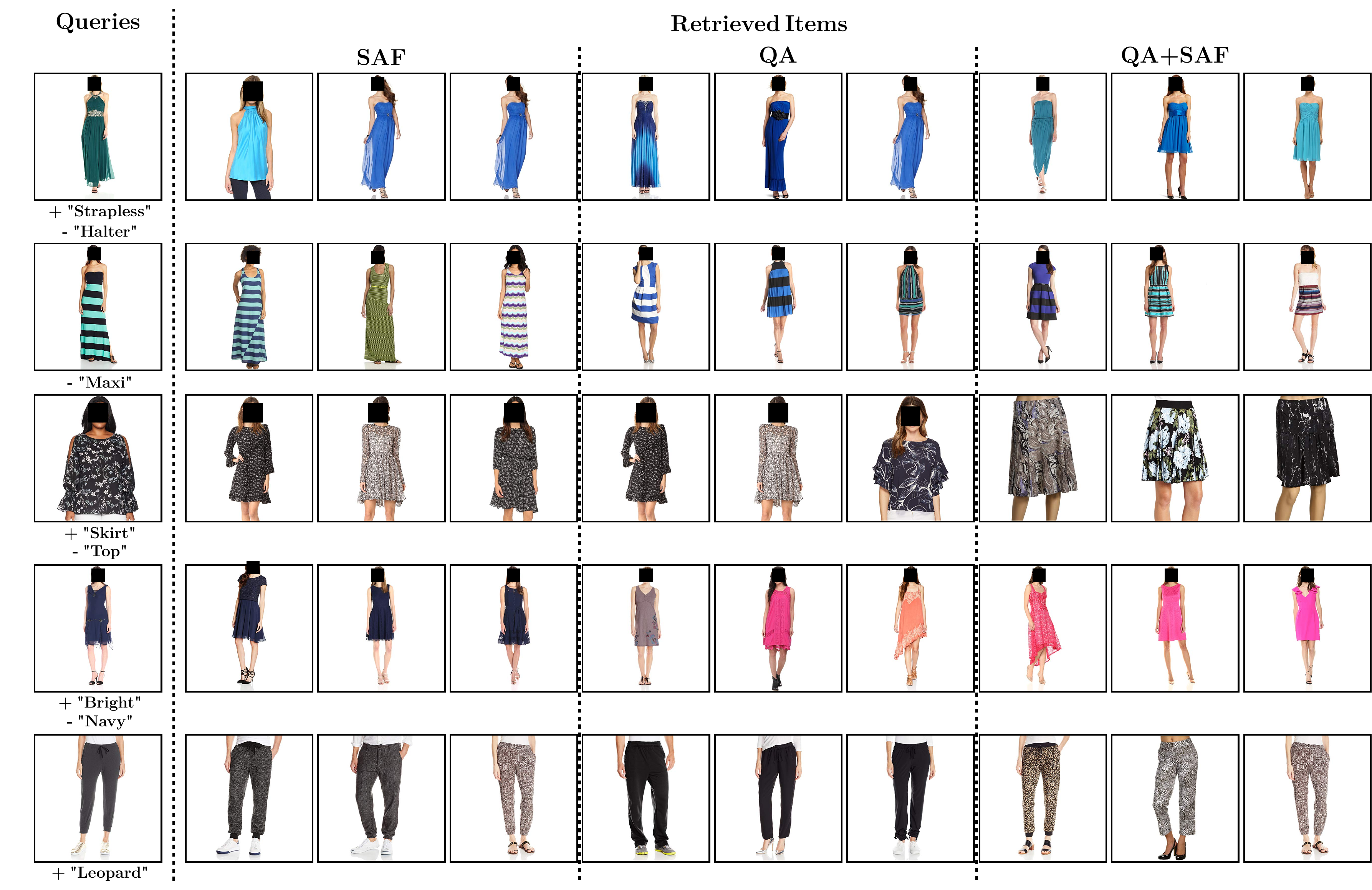}
	\caption{Qualitative results of top 3 retrieved items for example queries with Soft Attribute Filtering (SAF), Query Arithmetics (QA) and combined approach (QA+SAF).}
	
	\label{fig:results-examples}
\end{figure}

We follow our evaluation protocol for multimodal search, as described in Section~\ref{eval}, and compare the following methods: Soft Attribute Filtering (SAF), Query Arithmetic (QA) and their combination (QA+SAF). 
It can be seen in Table \ref{table:metric-eval} that although there is a clear trade-off between the visual and textual metrics (V-nDCG and T-nDCG), on the overall multimodal (MM) metric, the combined approach (QA+SAF) outperforms all others significantly. 
These conclusions are further reinforced by our qualitative visualization and analysis of the results, as can be seen in Figure \ref{fig:results-examples}.

\begin{table}[t]
	
	\begin{center}
		\bgroup
		\tiny
		\def\arraystretch{1.2}
		
		\caption{Evaluation results: we report V-nDCG, T-nDCG and MM metrics, and compare the Soft Attribute Filtering (SAF), the naive linear Query-Arithmetic (QA) and the combined (QA+SAF) methods.
			Queries are split by the type of textual criteria.}
		
		\label{table:metric-eval}
		
		\begin{tabular}{|l| |c|c|c|c|c|c|c|c|c|}

			\hline
			\multirow{2}{*}{} & \multicolumn{3}{|c|}{\textbf{SAF}} & \multicolumn{3}{|c|}{\textbf{QA}} & \multicolumn{3}{|c|}{\textbf{QA+SAF}}  \\
			
			\cline{2-10}
			& V-nDCG &T-nDCG & MM &  V-nDCG &T-nDCG & MM  &  V-nDCG &T-nDCG & MM  \\
			
			\hline
			\hline
			\cline{1-10}
			\textbf{Color}            &0.726 &0.407 &0.543  & 0.8 &0.413 &0.574  &0.621 &0.591 &\textbf{0.605}    \\
			\cline{1-10}
			\textbf{Pattern}           &0.769 &0.407  &0.559 &0.818    &0.426  &0.59  &0.672 &0.543  &\textbf{0.604}  \\
			\cline{1-10}
			\textbf{Neckline}            &0.77 &0.572  &\textbf{0.663} &0.806  &0.527  &0.651 &0.68 &0.628  &0.653  \\
			\cline{1-10}
			\textbf{Style}       		    &0.761 &0.464  &0.594 &0.815    &0.401  &0.572   &0.676 &0.563  &\textbf{0.617} \\
			\cline{1-10}
			\textbf{Garment}       		 &0.785   &0.27  &0.46 &0.828 &0.221 &0.427 &0.696 &0.486  &\textbf{0.581} \\
			\cline{1-10}
			\hline 		
			\hline
			\textbf{Overall}       		 &0.76    &0.419   &0.564  &0.813 &0.397 &0.568  &0.669 &0.561  &\textbf{0.612}   \\
			\hline		
		\end{tabular}
		\egroup
		
	\end{center} 
	
\end{table}


\section{Conclusions}
\label{sec:conclusion}

In this paper, we explored the task of multimodal fashion search. 
We proposed utilizing a visual-textual joint embedding model for this task, suggested an alternative training objective and demonstrated its effectiveness.
We explored and evaluated several approaches to leverage this joint-embedding model for the multimodal search task.
Unlike previous works, our method does not require direct supervised data of images before and after the textual manipulation.
Moreover, our training and evaluation methods are all performed over noisy, not well structured, catalog data.

\newpage

\medskip

\small
\bibliographystyle{plain}

\bibliography{all_bib}
\end{document}